\documentclass{article}

\usepackage{amsmath}
\usepackage{subcaption}

\usepackage{microtype}
\usepackage{graphicx}
\usepackage{booktabs} 

\usepackage{colortbl}
\usepackage{xcolor}
\usepackage{xspace}

\usepackage{siunitx}
\usepackage{multirow}

\usepackage{hyperref}

\newcommand{\mbs}{\unit{MB/s~}}
\newcommand{\highperf}{\underline}
\newcommand{\bestperf}[1]{\underline{\textbf{#1}}}
\newcommand{\geotiff}{\texttt{GeoTIFF}\xspace}
\newcommand{\zarr}{\texttt{Zarr}\xspace}
\newcommand{\deflate}{\texttt{DEFLATE}\xspace}
\newcommand{\lzw}{\texttt{LZW}\xspace}
\newcommand{\lz}{\texttt{LERC-ZSTD}\xspace}
\newcommand{\dask}{\texttt{Dask}\xspace}
\newcommand{\dl}{\texttt{DataLoader}\xspace}
\newcommand{\pt}{\texttt{PyTorch}\xspace}
\newcommand{\none}{\texttt{None}\xspace}
\newcommand{\deflateo}{\texttt{DEFLATE\_1}\xspace}
\newcommand{\deflates}{\texttt{DEFLATE\_6}\xspace}
\newcommand{\deflaten}{\texttt{DEFLATE\_9}\xspace}
\newcommand{\blocked}{\texttt{blocked}\xspace}
\newcommand{\threads}{\texttt{num\_threads}\xspace}
\newcommand{\patch}{\texttt{patch\_size}\xspace}
\newcommand{\workers}{\texttt{num\_workers}\xspace}
\newcommand{\prefetch}{\texttt{prefetch\_factor}\xspace}
\newcommand{\true}{\texttt{True}\xspace}
\newcommand{\false}{\texttt{False}\xspace}

\usepackage[accepted]{icml2025}

\usepackage{amsmath}
\usepackage{amssymb}
\usepackage{mathtools}
\usepackage{amsthm}

\usepackage[capitalize,noabbrev]{cleveref}

\theoremstyle{plain}

\theoremstyle{definition}

\theoremstyle{remark}

\usepackage[textsize=tiny]{todonotes}

\icmltitlerunning{Optimizing Cloud-to-GPU Throughput}

\begin{document}

\twocolumn[
\icmltitle{Optimizing Cloud-to-GPU Throughput for Deep Learning With Earth Observation Data}



\icmlsetsymbol{equal}{*}

\begin{icmlauthorlist}
\icmlauthor{Akram Zaytar}{equal,ai4g}
\icmlauthor{Caleb Robinson}{equal,ai4g}
\icmlauthor{Girmaw Abebe Tadesse}{ai4g}
\icmlauthor{Gilles Hacheme}{ai4g}
\icmlauthor{Tammy Glazer}{ai4g}
\icmlauthor{Anthony Ortiz}{ai4g}
\icmlauthor{Rahul Dodhia}{ai4g}
\icmlauthor{Juan M. Lavista Ferres}{ai4g}
\end{icmlauthorlist}

\icmlaffiliation{ai4g}{Microsoft AI for Good Research Lab, Redmond WA, USA}

\icmlcorrespondingauthor{Akram Zaytar}{akramzaytar@microsoft.com}

\icmlkeywords{Geospatial Deep Learning, Cloud-Native Training, Data Loader Optimization}

\vskip 0.3in
]



\printAffiliationsAndNotice{\icmlEqualContribution}

\begin{abstract}
Training deep learning models on petabyte-scale Earth observation (EO) data requires separating compute resources from data storage. However, standard PyTorch data loaders cannot keep modern GPUs utilized when streaming \geotiff files directly from cloud storage. In this work, we benchmark \geotiff loading throughput from both cloud object storage and local SSD, systematically testing different loader configurations and data parameters. We focus on tile-aligned reads and worker thread pools, using Bayesian optimization to find optimal settings for each storage type. Our optimized configurations increase remote data loading throughput by 20$\times$ and local throughput by 4$\times$ compared to default settings. On three public EO benchmarks, models trained with optimized remote loading achieve the same accuracy as local training within identical time budgets. We improve validation IoU by $6$--$15$\% and maintain $85$--$95$\% GPU utilization versus $0$--$30$\% with standard configurations. Code is publicly available at
\url{https://github.com/microsoft/pytorch-cloud-geotiff-optimization}
\end{abstract}

\section{Introduction}\label{introduction}
EO datasets have reached petabyte scale volumes~\cite{wilkinson2024environmental}, yet training deep learning models on this data faces a fundamental bottleneck. When streaming data directly from cloud storage, multiple factors introduce latency that leaves GPUs severely underutilized. The alternative of downloading entire datasets locally is impractical at this scale, as it demands significant storage capacity and repeated downloads for each experiment.

This bottleneck stems from both data loading mechanisms and format characteristics. The \pt \dl \cite{pytorch} was originally designed for local training scenarios with minimal access latency. Although it offers configuration options like worker processes, batch sizes, and pre-fetching to improve throughput, these assume fast local storage. At the same time, data format choices significantly impact remote access performance. Factors such as tiling block size, compression type and level, and image dimensions all affect how efficiently patches can be extracted from remote files. When data resides in cloud storage, the combination of suboptimal loader configurations and format characteristics can consume over half of each training epoch \cite{mohan}, leaving GPUs idle while waiting for the next batch.

The machine learning community has developed several approaches to address remote data loading challenges. These solutions generally follow three patterns: custom file formats for efficient streaming, asynchronous loading mechanisms, and shared caching systems. FFCV~\cite{ffcv} combines efficient file formats with asynchronous transfers to maximize GPU utilization. WebDataset~\cite{webdataset} uses tar archives to enable sequential reads from cloud storage, achieving ten-fold improvements over traditional approaches. NVIDIA DALI~\cite{nvidia_dali} offloads decoding and augmentation to GPUs to eliminate preprocessing bottlenecks. More recently, systems like Concurrent Dataloader~\cite{concurrent} and Amazon's S3 Connector~\cite{s3connector2023} enable concurrent fetching within batches, delivering up to 12$\times$ faster loading from cloud storage. Coordinated caching systems like CoorDL~\cite{coordl} share data across nodes to avoid redundant reads, yielding up to 15$\times$ speedup in multi-server settings.

\begin{figure*}[t]
    \centering
    \includegraphics[width=0.9\linewidth]{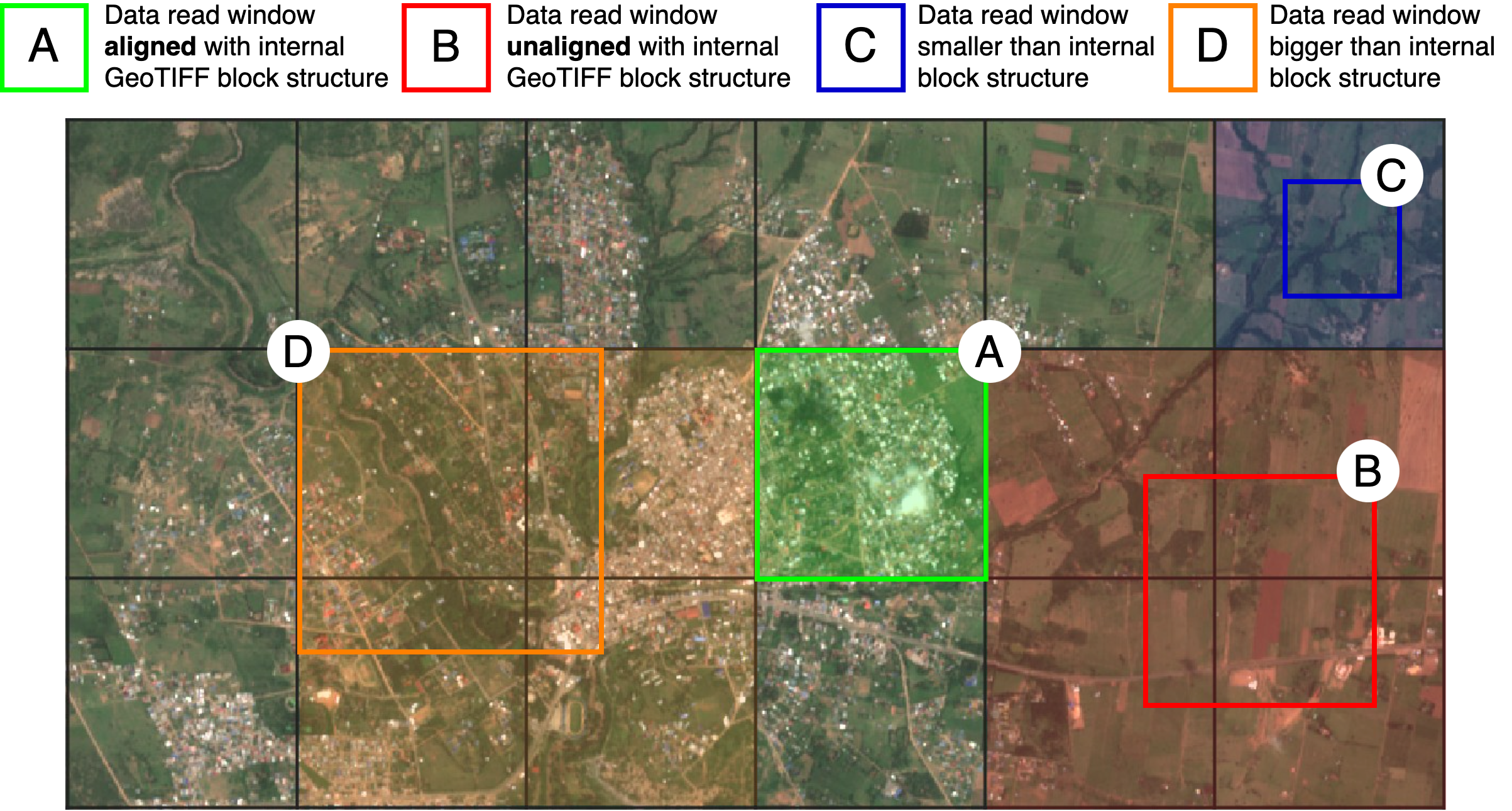}
    \caption{Sentinel-2 image with internal \geotiff block structure overlaid as black grid squares. Colored rectangles show four windowed read patterns and their data loading behavior (colored backgrounds). (\textcolor{green}{\textbf{A}}) shows an aligned read that matches block boundaries, efficiently loading only the requested data from a single block. (\textcolor{red}{\textbf{B}}) demonstrates a random window read of block size that intersects four blocks, requiring 4$\times$ more data transfer than needed. (\textcolor{blue}{\textbf{C}}) represents sub-block reads that still force entire block loading, creating overhead for small requests. (\textcolor{orange}{\textbf{D}}) shows large reads spanning multiple blocks, necessitating loading of all intersected blocks.}
    \label{fig:tiles}
\end{figure*}

While general approaches address key remote loading challenges, raster data presents unique characteristics requiring specialized solutions. Raster data is typically stored in large compressed tiles exceeding $10{,}000 \times 10{,}000$ pixels, from which training patches must be extracted through windowed requests or offline preprocessing. EO workflows have adopted cloud-native formats like Cloud Optimized \geotiff{}s (COGs)~\cite{cogeo2025} and \zarr{}~\cite{open2023zarr} arrays that enable efficient remote access through internal tiling schemes. In both formats, imagery is chunked into fixed $k\times k$ pixel tiles (Figure~\ref{fig:tiles}). An aligned window read is served by loading and decompressing exactly one tile, whereas a mis-aligned window read forces multiple tiles to be fetched and decompressed, inflating I/O and CPU costs. \geotiff{} supports various compression methods like \deflate{}, \lz{}, and \lzw{} that trade network transfer speed against CPU decompression overhead~\cite{alberti2023geotiff}. \zarr{} arrays, when combined with optimized \pt{}~\dl{}s and \dask{} for concurrent chunk loading, can achieve 15$\times$ throughput improvements while saturating GPU utilization~\cite{zarrxarray}. Additionally, \textit{xbatcher}~\cite{jones2023xbatcher} provides an interface for feeding \textit{XArray} data directly to \pt \dl{}s through batch generators that efficiently stream \zarr{} datasets from cloud storage.

However, systematic optimization for \geotiff workflows remains unexplored, despite most EO archives storing data in this format. While streaming optimization for \zarr \cite{open2023zarr} has been well-studied~\cite{zarrxarray}, \geotiff poses unique challenges due to significant variability in internal structure, compression methods, and tiling schemes across different archives. Classical approaches are ineffective for \geotiff workflows. Caching systems fail because geospatial sampling exhibits negligible data reuse--a consequence of vast spatio-temporal domains and diverse sampling strategies (random or spatially distributed) commonly used in EO model training. Custom file formats are also impractical due to conversion limitations when using COGs, though asynchronous loading remains applicable.

This paper addresses the gap through systematic optimization of both \pt{}~\dl{} configurations and \geotiff{} format characteristics specifically for remote streaming. Our contributions include:

\begin{itemize}
\item A benchmarking framework using Bayesian optimization to identify optimal configurations for \geotiff{} loading, focusing on tile-aligned reads and worker thread pools.
\item An optimized configuration achieving 20$\times$ higher remote throughput over baseline settings and 4$\times$ improvement for local reads.
\item Empirical validation through EO segmentation benchmarks showing optimized remote loading matches local-disk training accuracy within a fixed time budget.
\end{itemize}

\section{Experimental setup} \label{experimental_setup}
\subsection{Data preparation} \label{data_preparation}

Our goal is to simulate the real-world scenario of streaming raster patches from COGs hosted in Azure Cloud Storage. Since replicating an entire online archive for different compression types and levels is impractical, we address this by deliberately disabling caching to simulate fresh patch fetching and selecting a subset of images for our transforms and experimentation. We use Sentinel-2~\cite{drusch2012sentinel} multispectral satellite imagery, which provides global coverage at $10$~meter resolution with a $5$~day revisit time. This imagery is widely used in environmental monitoring~\cite{Gorrono2023}, agriculture~\cite{Segarra2020}, and land cover classification tasks~\cite{Phiri2020}, making it representative of typical EO deep learning workflows.

We download data through Microsoft's Planetary Computer SpatioTemporal Asset Catalogs API~\cite{microsoft_open_source_2022_7261897}, focusing on a 168 km$^2$ region centered on Nairobi, Kenya. We selected imagery from $2024$ with cloud cover below 5\%, resulting in $10$ scenes for our benchmark dataset. From each scene, we extract four spectral bands (``B02'', ``B03'', ``B04'', and ``B08''), representing blue, green, red, and near-infrared, respectively.

To systematically evaluate how compression affects data loading throughput, we process each scene into $6$ compression variants: uncompressed (\none), \lzw with predictor $2$, \deflate at three different levels (i.e, \deflateo, \deflates, and \deflaten), and \lz. These methods represent different compression philosophies: \lzw and \deflate provide varying speed-compression tradeoffs (where higher \deflate levels achieve better compression at the cost of slower encoding), while \lz offers an alternative approach optimized for numerical data. This allowed us to analyze the tradeoffs between file size and processing speed across various compression methods.

We store all variants as COGs with consistent internal tiling structure ($512 \times 512$ pixel blocks). The resulting dataset consists of $60$ GeoTIFF files ($10$ scenes $\times$ $6$ compression variants). We upload all dataset variants to Azure Blob Storage while maintaining an identical local copy for comparative benchmarking. This setup enables direct comparison between cloud and local storage performance across different compression configurations.

\subsection{Compute environment} \label{infrastructure}
We conduct all experiments on an Azure \texttt{Standard\_NC96ads\_A100\_v4} instance, featuring 96 AMD EPYC 7V13 vCPUs, 866~GB RAM, and an \texttt{~NVIDIA~A100~80GB} GPU in the West US 3 region. For local storage benchmarks, we use Azure temporary storage with a single \texttt{250~GB} disk formatted with the \texttt{ext4} filesystem. For remote storage experiments, we use Azure Blob Storage with \texttt{Standard\_LRS} in the \texttt{Hot} tier, located in West US 2, resulting in approximately \texttt{164~ms} cross-region latency.

\section{Methods}
\label{methods}

We aim to maximize the data loading throughput delivered to the training loop under different hyperparameter configurations. Throughput is measured in megabytes per second (\mbs) and represents how much image data the pipeline can process per unit time. To calculate throughput, we time the complete data loading process. We count the total number of processed pixels and convert it to bytes. The throughput is then computed as the total data volume processed divided by the elapsed wall-clock time. This measurement captures the end-to-end performance of the data loading pipeline.

\subsection{Internal tile sampling}\label{subsec:internal_tile_sampling}

We introduce a binary hyperparameter $\blocked \in\!\{True,False\}$ that enforces read alignment in the \dl. When $\blocked=\true$, the sampler chooses a random $k{\times}k$ tile, then--if the window size is $p{\times}p$ and $p\le k$--we jitter the window to a random position that stays wholly inside that tile (\texttt{(C)} in Figure~\ref{fig:tiles}); otherwise the window starts at the tile origin. Any window that fits in one tile is thus served by a single block read, cutting I/O by up to $4{\times}$.

\subsection{Worker thread pools}\label{subsec:thread_pools}

We introduce an intra-worker thread pool of width \threads $\in\{1,2,\dots,32\}$. Each worker issues up to \texttt{num\_threads} concurrent range requests, hiding the $\approx$164~ms round-trip latency of Azure Blob Storage behind computation already in flight. This approach aims to increase throughput without spawning more workers.

\subsection{Bayesian search}\label{subsec:bo}
The search space is in the order of $10^{4}$ configurations, making exhaustive search impractical. We employ Optuna~\cite{optuna_2019} with the Tree-structured Parzen Estimator~\cite{bergstra2011algorithms}, which builds probability models of good and bad configurations to select candidates with high expected improvement. We explore: compression, patch size, number of workers, thread pool size,  block alignment and batches preloaded per worker (prefetch factor) as shown in Table~\ref{tab:data_loading_configs}. Each trial measures throughput (\mbs) over $5$ epochs, with each epoch streaming $1024$ patches. We conduct $100$ trials without early stopping criteria, monitoring both throughput and average GPU utilization.

\begin{table}[htbp]
\centering
\caption{Experimental configurations explored in the data loading pipeline.}
\begin{tabular}{l p{3cm}}  
\toprule
\textbf{Parameter} & \textbf{Values} \\
\midrule
Compression & \none, \texttt{deflate}\{1,6,9\}, \texttt{lzw}, \texttt{lerc-zstd} \\
\patch & $\{128, 256, 512, 1024\}$ \\
\workers ($w$) & $\{1, 2, 4, 8, 16, 32, 64\}$ \\
\threads & $\{1, 2, 4, 8, 16\}$ \\
\blocked &  $\{\true, \false\}$  \\
\prefetch & $\{1, 2, 4, 8, 16\}$ \\
\bottomrule
\end{tabular}
\label{tab:data_loading_configs}
\end{table}

\subsection{Grid search}\label{subsec:grid_search}
While Bayesian optimization quickly finds promising regions in the hyperparameter space, it may miss important interactions between specific hyperparameter pairs. To address this, we implement $2D$ grid search that explores the cross-product of values for selected hyperparameter pairs. For each combination, we run $5$ data loading simulations and calculate the mean and standard deviation of throughput.

We focus our grid search on hyperparameter pairs with the highest feature importance scores derived from our Bayesian optimization results, maintaining consistent experimental conditions and setting optimal values for the fixed hyperparameters. This approach creates detailed visualizations of the performance landscape that reveal how different hyperparameter combinations influence data loading performance.

\section{Results}
\label{results}

Our Bayesian optimization search revealed distinct optimal configurations for local versus remote storage scenarios, as shown in Table~\ref{tab:bayesian_results}. Local storage achieved peak throughput of $1285 \pm 29$ \mbs with uncompressed imagery and block-aligned patch size reads, while remote storage reached $849 \pm 51$ \mbs using $64$ workers and \lz compression. This significant difference in optimal configurations highlights the need for storage-specific optimization approaches. Feature importance in Table~\ref{tab:feature_importance} tells the same story, showing different optimization priorities between storage types. For local storage, compression type (35.24\%) and \threads (32.46\%) dominate performance factors. In contrast, remote storage performance is primarily determined by \workers (28.08\%) and \threads (26.06\%), with compression type contributing only 8.59\% to performance variation.

\begin{table*}[t]
\caption{Bayesian optimization search: throughput improvement over baseline configurations. \textbf{Optimization achieved dramatic speedups} of 4.1$\times$ for local storage (1285 vs 313 \mbs) and 20.5$\times$ for remote storage (849 vs 41 \mbs), with remote storage benefiting more from worker scaling and local storage from eliminating compression overhead.}
\label{tab:bayesian_results}
\vskip 0.15in
\begin{center}
\begin{small}
\begin{sc}
\begin{tabular}{lccccccc}
\toprule
Storage & Compression & \patch & \blocked & \workers & \threads & Throughput & Speedup \\
\midrule
\multirow{2}{*}{Local}  & \deflates & 256  & \false & 4  & 1 & 313  $\pm$ 12 \mbs & - \\
                       & \none     & 512  & \true  & 4  & 1 & 1285 $\pm$ 29 \mbs & 4.1x \\
\midrule
\multirow{2}{*}{Remote} & \deflates & 256  & \false & 4  & 1 & 41   $\pm$ 1  \mbs & - \\
                       & \lz       & 1024 & \true  & 64 & 1 & 849  $\pm$ 51 \mbs & 20.5x \\
\bottomrule
\end{tabular}
\end{sc}
\end{small}
\end{center}
\vskip -0.1in
\end{table*}

\begin{table*}[t]
\caption{Feature-importance rankings for the hyperparameters. Importance scores calculated using functional ANOVA with a random forest surrogate model to decompose variance in the objective function attributable to each hyperparameter. Parameter importance varies significantly between storage types: local performance depends primarily on compression (35\%) while remote performance is dominated by worker count (28\%), reflecting different bottlenecks in each scenario.}
\label{tab:feature_importance}
\vskip 0.15in
\begin{center}
\begin{small}
\begin{sc}
\begin{tabular}{r l c l c}
\toprule
& \multicolumn{2}{c}{Local} & \multicolumn{2}{c}{Remote} \\ 
\cmidrule(lr){2-3} \cmidrule(lr){4-5}
Rank & Parameter & Importance (\%) & Parameter & Importance (\%) \\
\midrule
1 & compression & 35.24 & \texttt{num\_workers} & 28.08 \\
2 & \threads & 32.46 & \texttt{num\_threads} & 26.06 \\
3 & \patch & 25.63 & \texttt{patch\_size} & 17.95 \\
4 & \workers & 4.22 & \texttt{Blocked} & 11.11 \\
5 & \prefetch & 1.97 & compression & 8.59 \\
\bottomrule
\end{tabular}
\end{sc}
\end{small}
\end{center}
\vskip -0.1in
\end{table*}

Compression strategies showed context-dependent effectiveness. As Table~\ref{tab:throughput} demonstrates, uncompressed data consistently outperformed all compression methods for local storage, with throughput advantages of 1.3-1.6$\times$ (974 \mbs for LERC-ZSTD vs.\ 1286 \mbs for uncompressed at optimal worker counts). For remote storage, LERC-ZSTD compression balanced network transfer efficiency with reasonable decompression overhead, making it the optimal choice despite compression being less important overall.

Tables \ref{tab:patch_workers} and \ref{tab:sampler_performance} consistently show maximum throughput at 1024-pixel patches across both local and remote storage. Notably, this value represents our hyperparameter search's upper bound, suggesting potential for further optimization (although limited by GPU memory). The 1024-pixel patches align efficiently with model processing by fetching $2 \times 2$ tiles of $512$ pixels each, enabling complete utilization without wasted computation.

Worker-thread interactions showed counter-intuitive results in Table~\ref{tab:workers_threads}. Multi-worker setups with minimal threading achieved higher throughput. Two key limitations likely account for these results: cloud provider rate-limiting when too many worker-thread concurrent requests are made, and our thread pool implementation requiring all requests to complete before returning data, causing a single slow request to degrade overall performance.

For remote storage, block-aligned sampling substantially outperformed random access. Table~\ref{tab:sampler_performance} shows this advantage growing dramatically with patch size--from 45\% improvement at 128 pixels (32 \mbs vs.\ 22 \mbs) to 79\% at 1024 pixels (810 \mbs vs.\ 453 \mbs). This confirms that respecting tile boundaries significantly reduces unnecessary data transfers, with benefits compounding at larger scales. Remote storage loading also benefited from prefetching, converging on factor of $8$ to mask the higher network latency. This difference reflects the fundamental need to hide connection latency when streaming from cloud storage.

Based on our findings, we recommend the following practices for training on cloud raster imagery: \textbf{match the patch size to the underlying tile structure} of your \geotiff  files (typically $256$ or $512$ pixels), \textbf{use block-aligned sampling} to prevent inefficient partial tile reads, \textbf{prioritize high worker counts} ($32$--$64$) over threads, \textbf{consider \lz compression} to balance transfer and decompression, and \textbf{implement aggressive pre-fetching} (factor of $8$).

\section{Impact of optimized loading on training}
\label{impact_on_training}

\begin{figure*}[t]
    \centering
    \includegraphics[width=\linewidth]{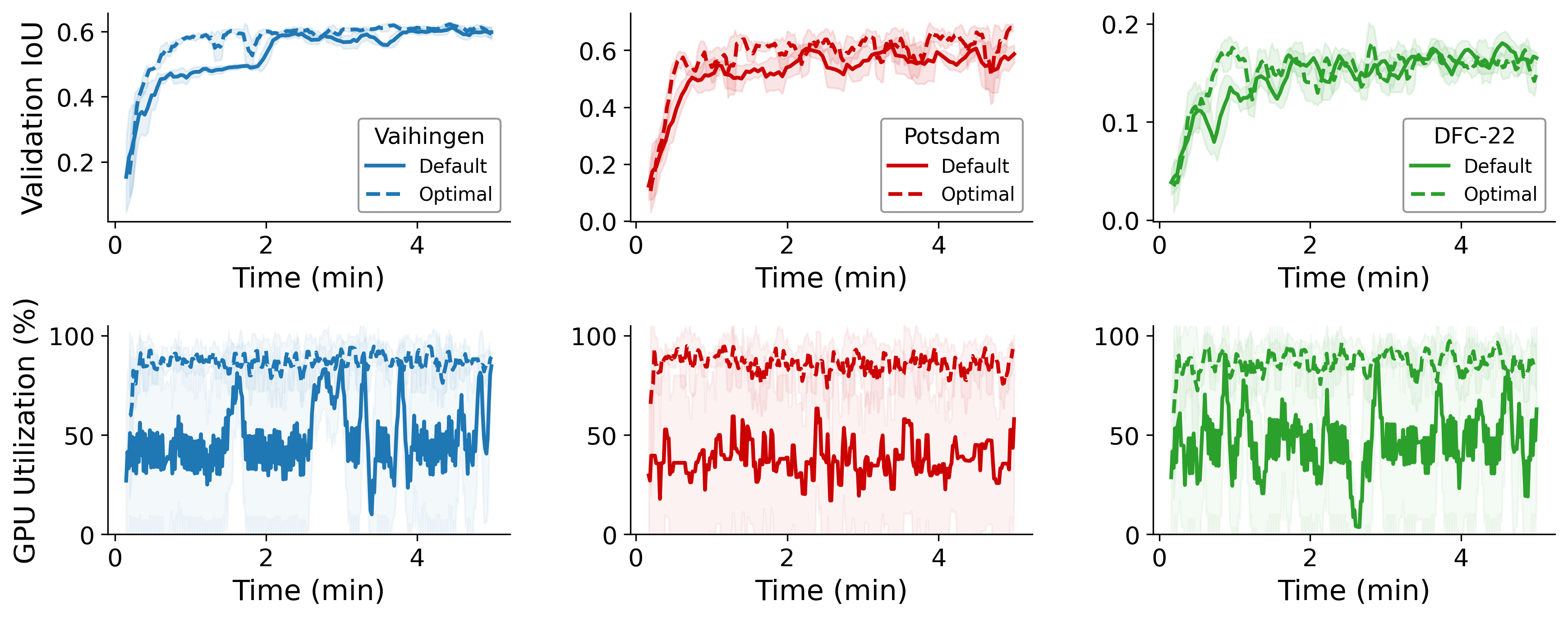}
    \caption{\textbf{Training on SSD.} Top row: Validation IoU over time shows optimal configurations (dashed lines) achieving marginal improvements over default configurations (solid lines) for Vaihingen (blue, $0.6$ vs $0.5$), while reaching similar final performance for Potsdam (red, $0.6$) and DFC-22 (green, $0.17$) datasets. Bottom row: GPU utilization reveals the key advantage of optimal configurations, which maintain consistent high utilization ($90$\%) across all datasets compared to default configurations' variable utilization.}
    \label{fig:local_training_performance}
\end{figure*}

\begin{figure*}[t]
    \centering
    \includegraphics[width=\linewidth]{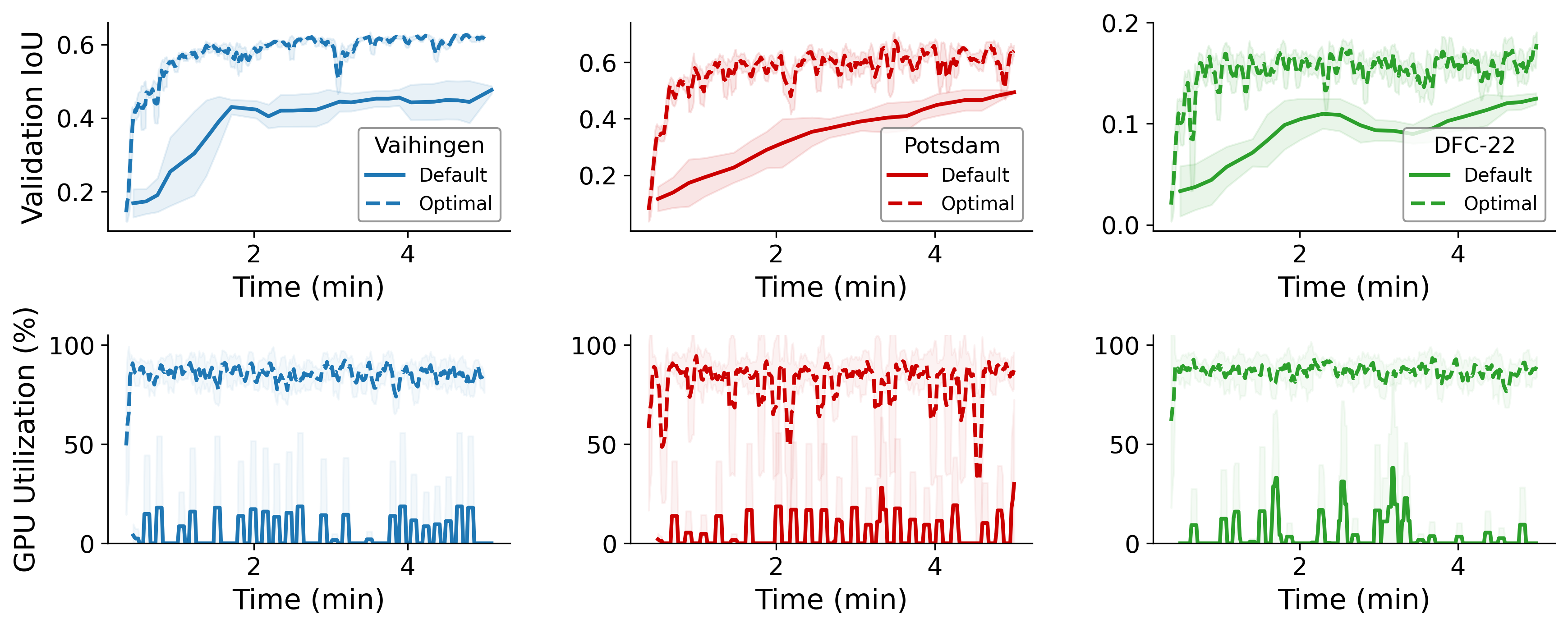}
    \caption{\textbf{Training on Azure Cloud Storage}. Top row: Validation IoU over time shows optimal configurations (dashed lines) consistently outperforming default ones (solid lines) across Vaihingen (blue, $0.6$ vs $0.45$), Potsdam (red, $0.6$ vs $0.5$), and DFC-22 (green, $0.18$ vs $0.12$) datasets. Bottom row: GPU utilization demonstrates sustained high usage ($80$--$90$\%) for optimal configurations versus intermittent, low utilization ($\le$ 30\%) for default settings.}
    \label{fig:remote_training_performance}
\end{figure*}

We evaluate loading configurations on model training using three standard semantic segmentation benchmarks. The ISPRS Vaihingen dataset~\cite{isprs_semantic_labeling_2014} contains 33 high-resolution (9~cm) aerial scenes of urban areas in Vaihingen, Germany, with three spectral bands (near-infrared, red, green) and digital surface models. The Potsdam dataset~\cite{isprs_semantic_labeling_2014} provides 38 tiles of $6000 \times 6000$ pixels at 5~cm resolution with four spectral bands (RGB plus near-infrared). Both datasets feature six manually-labeled land cover classes (impervious surfaces, buildings, low vegetation, trees, cars, and clutter/background). The IEEE GRSS DFC-22 dataset~\cite{hansch20222022} offers RGB imagery from 19 French urban areas at 50~cm resolution with approximately $2000 \times 2000$ pixel images, featuring 14 land cover classes for more complex segmentation tasks.

To systematically evaluate the impact of storage format and loader configuration on training performance, we create three dataset versions for each benchmark. The \texttt{default} version converts original \geotiff files to COGs using \deflate compression (\texttt{zlevel} 6) with $512 \times 512$ tiles. The \texttt{local-optimal} version uses uncompressed data, while \texttt{remote-optimal} applies \lz compression based on our Bayesian optimization results for cloud storage. We upload \texttt{default} and \texttt{remote-optimal} version copies to Azure Blob Storage to test remote training scenarios. Each dataset version uses storage-specific loader configurations: \texttt{default} employs standard PyTorch settings ($4$ workers, $256$-pixel patches), \texttt{local-optimal} uses $4$ workers with $512$-pixel patches, and \texttt{remote-optimal} uses $64$ workers with 8$\times$ prefetch factor and block-aligned reads.

We train a \textit{UNet} ~\cite{unet} segmentation model with a \textit{ResNet-18} ~\cite{he2016deep} encoder using \textit{AdamW} ~\cite{adamw} optimization with a learning rate of $10^{-3}$ and standard data augmentation techniques (flips, rotations). We employ a fixed $5$-minute training time budget for each configuration, as ``speed-runs'' are good for measuring how data loading efficiency impacts training dynamics~\cite{coleman2017dawnbench,mattson2020mlperf} before performance curves saturate. We measure validation IoU and GPU utilization throughout the process. GPU utilization was sampled at $1$-second intervals using NVIDIA Management Library in a separate thread running concurrently with training. To ensure fair comparison, we use identical model architecture, hyperparameters, and validation data across all configurations, isolating the impact of data loading strategies.

The results visualized in Figure~\ref{fig:local_training_performance} and Figure~\ref{fig:remote_training_performance} demonstrate striking differences between configurations. For local storage (Figure~\ref{fig:local_training_performance}, top row), both default and optimal configurations converged to similar validation IoU values for all three datasets, with the main difference being GPU utilization--the optimal configuration maintained significantly higher GPU usage (80-100\% vs. 30-60\%). The remote storage comparison (Figure~\ref{fig:remote_training_performance}) revealed a more substantial difference: the \texttt{remote-optimal} configuration achieved comparable performance to local storage with sustained high GPU utilization, while the \texttt{default} configuration struggled to reach comparable IoU values with highly erratic GPU usage, often dropping to 0\%. This represents a performance gap of approximately $15\%$ in model accuracy on Vaihingen dataset ($0.6$ vs $0.45$), 10\% on Potsdam dataset ($0.6$ vs $0.5$), and 6\% on DFC-22 dataset ($0.18$ vs $0.12$), attributable solely to data loading configuration.

These results highlight the critical importance of data format and loading configurations in EO deep learning. Properly aligned reads, optimal worker counts, and appropriate compression formats directly translate to higher GPU utilization and better performance. Most importantly, our experiments demonstrate that with optimized configurations, models can train directly on cloud-stored data without performance penalty, eliminating the need for costly local storage.

\section{Limitations}
\label{limitations}

Our approach has several practical limitations. Public archives like Microsoft Planetary Computer use fixed data formats that cannot be modified, restricting our optimizations to data loading rather than storage format improvements. While we could transform and re-store data on-the-fly, this would add infrastructure complexity. Some optimization choices may negatively impact deep learning workflows. For example, tile-aligned reads prevent random cropping during training, potentially reducing model performance. Our investigation also focuses exclusively on lossless compression methods within standard \pt \dl configurations. We do not explore lossy compression tradeoffs, which could offer different performance characteristics depending on imagery spatial resolution and downstream tasks. Additionally, we do not account for GDAL environment variables that can affect read performance. Finally, our evaluation assumes remote data access across different data centers (Azure West US 2 and 3). Co-locating data and compute would reduce latency and may change our optimization recommendations.

\section{Future work}
\label{future_work}

As cloud-hosted EO datasets continue to grow in scale, traditional epoch-based training becomes increasingly impractical. Hence, we propose replacing epochs with infinite samplers that continuously fetch data patches using asynchronous, non-blocking frameworks. This eliminates artificial epoch boundaries and samples according to defined probability distributions. These training systems must handle network failures gracefully through retry mechanisms, placeholder tensors, and data replication to maintain batch sizes when network issues occur. Beyond training, we need systematic studies comparing streaming efficiency across data formats beyond COGs, including ZARR, covering chunking strategies, compression methods, and access patterns for each format. Finally, the EO community should establish standardized performance benchmarks similar to \textit{DAWNBench}~\cite{coleman2017dawnbench} and \textit{MLPerf} \cite{mattson2020mlperf} to drive discovery of efficient training practices for cloud imagery and enable fair comparisons across methods.

\section*{Impact statement}
\label{impact}
Our work addresses the efficiency bottlenecks in EO deep learning by optimizing Cloud-to-GPU data streaming. Our approach removes a key barrier to experimentation, enabling faster iteration in geospatial machine learning. As EO data volumes continue growing and geospatial foundation models emerge, efficient data streaming becomes critical for scalable model development. Our optimizations reduce computational waste, lower development costs, and enable more efficient resource utilization across the research community. The primary societal benefit is faster insight generation for environmental monitoring and decision making. Researchers can now train models directly on expanding satellite archives, facilitating timely incorporation of new observations for dynamic systems like vegetation change, urban development, and natural disasters. This improved accessibility has the potential to accelerate progress on pressing environmental challenges.

\bibliography{paper}
\bibliographystyle{icml2025}

\newpage
\appendix
\onecolumn

\section{Grid search results} \label{appendix:grid_results}
\setlength{\intextsep}{6pt}

\begin{table}[H]
\small
\caption{\textbf{Local Throughput}: Num Workers vs. Compression. \textbf{Uncompressed data consistently outperforms all compression methods} by 1.3-1.6$\times$, achieving peak performance of 1286 \mbs with 8 workers, demonstrating that compression overhead outweighs benefits for local storage access. Values: \mbs $\pm$ std. (underlined: column best; bold underlined: overall best)}
\label{tab:throughput}
\vskip 0.1in
\begin{center}
\begin{small}
\begin{sc}
\begin{tabular}{r c c c c c c}
\toprule
& \multicolumn{6}{c}{Compression} \\
\cmidrule(lr){2-7}
\workers & \texttt{DEFLATE\_1} & \texttt{DEFLATE\_6} & \texttt{DEFLATE\_9} & \texttt{\lz} & \texttt{\lzw} & \none \\
\midrule
1 & 268 (11) & 268 (15) & 270 (03) & 388 (08) & 295 (07) & 610 (05) \\
2 & 491 (17) & 496 (02) & 475 (04) & 685 (13) & 525 (21) & 981 (17) \\
4 & 764 (42) & 748 (19) & 746 (14) & \highperf{974} (14) & 806 (67) & 1244 (48) \\
8 & \highperf{786} (18) & \highperf{761} (12) & \highperf{747} (16) & 867 (61) & 796 (40) & \bestperf{1286} (22) \\
16 & 782 (22) & 753 (16) & 720 (11) & 856 (53) & \highperf{818} (45) & 1242 (26) \\
32 & 765 (45) & 750 (20) & 730 (22) & 876 (73) & 773 (50) & 1200 (33) \\
64 & 705 (71) & 719 (29) & 711 (14) & 857 (101) & 690 (61) & 1184 (14) \\
\bottomrule
\end{tabular}
\end{sc}
\end{small}
\end{center}
\end{table}

\begin{table}[H]
\small
\caption{\textbf{Local Throughput}: Num. Workers vs. Patch Size. \textbf{Larger patch sizes consistently yield higher throughput}, with 1024-pixel patches achieving optimal performance (1298 \mbs at 8 workers) by efficiently utilizing GPU memory through $2 \times 2$ tiles of 512 pixels each. Values: \mbs $\pm$ std. (underlined: column best; bold underlined: overall best)}
\label{tab:patch_workers}
\vskip 0.1in
\begin{center}
\begin{small}
\begin{sc}
\begin{tabular}{r c c c c}
\toprule
& \multicolumn{4}{c}{\patch} \\
\cmidrule(lr){2-5}
\workers & 128 & 256 & 512 & 1024 \\
\midrule
1 & 127 (06) & 297 (07) & 592 (12) & 674 (31) \\
2 & 229 (05) & 510 (15) & 946 (09) & 1058 (20) \\
4 & \highperf{340} (18) & 701 (31) & 1147 (40) & 1269 (40) \\
8 & 323 (17) & \highperf{738} (40) & 1124 (33) & \bestperf{1298} (26) \\
16 & 310 (08) & 719 (16) & \highperf{1148} (08) & 1297 (37) \\
32 & 279 (11) & 682 (10) & 1105 (24) & 1273 (09) \\
64 & 211 (10) & 617 (10) & 1055 (21) & 1217 (50) \\
\bottomrule
\end{tabular}
\end{sc}
\end{small}
\end{center}
\end{table}

\begin{table}[H]
\small
\caption{\textbf{Remote Throughput}: Sampler vs. Patch Size. \textbf{Block-aligned sampling dramatically outperforms random access}, with advantages growing from 45\% at 128 pixels to 79\% at 1024 pixels (810 vs 453 \mbs), confirming that respecting tile boundaries minimizes unnecessary data transfers. Values: \mbs $\pm$ std. (underlined: column best; bold underlined: overall best)}
\label{tab:sampler_performance}
\vskip 0.1in
\begin{center}
\begin{small}
\begin{sc}
\begin{tabular}{r c c c c}
\toprule
& \multicolumn{4}{c}{\patch} \\
\cmidrule(lr){2-5}
\texttt{Blocked} & 128 & 256 & 512 & 1024 \\
\midrule
\false & 22 (8) & 83 (20) & 254 (79) & 453 (81) \\
\true & \highperf{32} (1) & \highperf{117} (9) & \highperf{401} (4) & \bestperf{810} (57) \\
\bottomrule
\end{tabular}
\end{sc}
\end{small}
\end{center}
\end{table}

\begin{table}[H]
\small
\caption{\textbf{Remote Throughput}: Workers vs. Threads. Counter-intuitively, fewer threads with more workers achieve higher throughput (817 \mbs with 16 workers, 1 thread), likely due to cloud provider rate-limiting and thread pool synchronization bottlenecks that cause slow requests to degrade overall performance. Values: \mbs $\pm$ std. (underlined: column best; bold underlined: overall best)}
\label{tab:workers_threads}
\vskip 0.1in
\begin{center}
\begin{small}
\begin{sc}
\begin{tabular}{r c c c c}
\toprule
& \multicolumn{4}{c}{Number of  Threads} \\
\cmidrule(lr){2-5}
Work. & 1 & 2 & 8 & 32 \\
\midrule
1 & 112 (03) & 187 (16) & 359 (08) & 362 (19) \\
2 & 213 (10) & 340 (09) & 486 (38) & \highperf{402} (20) \\
4 & 412 (12) & 560 (17) & \highperf{497} (38) & 356 (41) \\
8 & 600 (52) & \highperf{729} (58) & 511 (11) & 372 (37) \\
16 & \bestperf{817} (61) & 669 (118) & 487 (30) & 375 (11) \\
32 & 773 (113) & 666 (119) & 496 (20) & 371 (27) \\
64 & 761 (180) & 686 (85) & 474 (60) & 381 (19) \\
\bottomrule
\end{tabular}
\end{sc}
\end{small}
\end{center}
\end{table}

\end{document}